\documentclass{article}

\PassOptionsToPackage{numbers, compress}{natbib}

\usepackage[preprint,main]{neurips_2026}

\usepackage[utf8]{inputenc} % allow utf-8 input
\usepackage[T1]{fontenc}    % use 8-bit T1 fonts
\usepackage{hyperref}       % hyperlinks
\usepackage{url}            % simple URL typesetting
\usepackage{booktabs}       % professional-quality tables
\usepackage{amsfonts}       % blackboard math symbols
\usepackage{nicefrac}       % compact symbols for 1/2, etc.
\usepackage{microtype}      % microtypography
\usepackage{xcolor}         % colors
\usepackage{amsmath}        % for math equations
\usepackage{graphicx}       % for figures
\usepackage{amssymb}
\usepackage{enumitem}       % for tight lists
\usepackage{wrapfig}
\usepackage{makecell}       % for table cell formatting
\usepackage{multirow}       % for multirow table headers
\usepackage{algorithm}
\usepackage{algpseudocode}
\usepackage{amsmath, amssymb, amsthm}

\title{Dual-Layer Agentic Memory with Fast Write Routing and Slow Consolidation}

\author{%
  Wenzhi Li \\
  Zhejiang University
  \And
  Dong Nie \\
  Independent Researcher
  \And
  Tongtong Lyu \\
  Xiaohongshu
  \And
  Rui Lan \\
  Xiaohongshu
  \And
  Peiyao Wang \\
  Independent Researcher
  \And
  Lingzi Hong \\
  University of North Texas
  \And
  Weihang Pan\thanks{Corresponding author. Email:
  \href{mailto:panweihang@zju.edu.cn}{panweihang@zju.edu.cn}} \\
  Zhejiang University
  \And
  Binbin Lin \\
  Zhejiang University
  \And
  Boyuan Pan \\
  Xiaohongshu
  \And
  Yao Hu \\
  Xiaohongshu
}

\begin{document}

\maketitle

\begin{abstract}

Large language model (LLM) agents operate in dynamic environments where knowledge continuously evolves. Existing memory systems typically treat external memory as a monotonically growing repository, inevitably leading to retrieval degradation and increasing computational costs over time. We argue that the core challenge is not retrieval alone, but managing the \emph{knowledge lifecycle}: deciding what to externalize, update, or ultimately internalize. Inspired by Complementary Learning Systems (CLS) theory in neuroscience, we propose \textbf{Dual-Layer Agentic Memory}, a framework that shifts memory management to the write phase through cost-aware epistemic routing and periodic parametric consolidation. Incoming information is categorized as \textit{non-write}, \textit{write-new}, or \textit{write-update}, and routed through a small-to-large model cascade that minimizes routing overhead while filtering redundant memories. A subsequent write-back phase selectively consolidates high-value external memories into model parameters via supervised fine-tuning. Experiments demonstrate the dual efficiency of our approach: a 1.7B/8B cascade prunes up to 68\% of redundant external memory while escalating fewer than 50\% of inputs, yet retains over 98\% of the downstream QA Exact Match (EM) achieved by an exhaustive retention baseline. We further show that periodic consolidation successfully internalizes external knowledge, allowing the router to adaptively suppress redundant writes as the model's epistemic boundaries evolve. Overall, our framework presents a unified paradigm for agent memory: selective externalization followed by selective internalization. Code and dataset will be released upon acceptance.
\end{abstract}

\section{Introduction}
\label{sec:intro}

Large language model (LLM) agents are increasingly deployed in long-horizon, dynamic environments such as multi-session dialogue, autonomous research, personal assistants, and continual decision-making systems. In these settings, knowledge evolves over time rather than remaining static. An agent may repeatedly encounter information that is already captured by its parameters, genuinely novel facts that lie beyond its current parametric memory, or updates that directly conflict with stale internal beliefs. For example, an assistant may accumulate many user interaction logs that are redundant or outdated, while failing to preserve a newly stated user preference or revise a previously incorrect assumption. As a result, effective memory management is essential for persistent and coherent reasoning in agent systems.

Most existing approaches tackle this challenge by equipping agents with explicit external memory mechanisms, typically implemented through retrieval-augmented generation (RAG) \cite{lewis2020retrieval}, memory logs \cite{park2023generative}, hierarchical buffers \cite{packer2023memgpt}, or structured external memory. These systems improve persistence beyond the context window, but they usually treat external memory as an unconditional log into which observations are monotonically appended \cite{park2023generative}. Because filtering mechanisms operate primarily in the \emph{read} phase—relying on downstream queries to condition retrieval—systems cannot proactively reject redundant information at the source. This design leads to a fundamental storage--performance trade-off: as external memory grows, retrieval becomes noisier, more expensive, and more latency-sensitive.

A different line of work attempts to bypass retrieval by directly incorporating new knowledge into model parameters through knowledge editing \cite{meng2022locating, meng2022mass} or continual learning \cite{wu2024continual}. While this enables fast retrieval-free access, it comes with its own limitations: parameter updates are computationally expensive and prone to interference or catastrophic forgetting. More importantly, both retrieval-heavy and parametric-only approaches implicitly assume a fixed storage substrate. Knowledge is either externalized or internalized, but is rarely treated as something that should be dynamically allocated, revised, and migrated across memory substrates over time.

We argue that this view is fundamentally incomplete. Rather than deferring filtering decisions to a query-driven read phase, the core challenge is managing knowledge across its full lifecycle the moment it enters the system. When a new fact arrives, the agent should immediately evaluate it against its parametric memory to decide whether the fact is redundant, whether it must be written into external memory, and whether it should later be consolidated into model parameters. In particular, external memory should not serve as a permanent dumping ground. Instead, it should function as a selective and temporary store for knowledge that is either missing from parametric memory or in conflict with stale internal beliefs.

This perspective is naturally connected to memory mechanisms in neuroscience. Complementary learning systems (CLS) theory \cite{kumaran2016learning} posits that biological memory relies on two interacting subsystems: a fast-learning hippocampus capable of rapidly encoding novel experiences, and a slower neocortex that gradually consolidates stable knowledge into long-term representations, typically via offline replay during sleep. From this perspective, external memory acts as the fast, editable hippocampal buffer for novel or conflict-prone information, while parametric memory serves as the slower, stable cortical substrate. Inspired by this analogy, we formulate agent memory as a \emph{dual-layer knowledge lifecycle} problem consisting of fast write routing and slow consolidation (analogous to sleep-dependent memory replay).

Concretely, we propose a dual-layer system with two types of memory and two forms of decisions. At the \textbf{carrier layer}, the agent maintains \emph{parametric memory} for stable, compressed, low-cost knowledge and \emph{external memory} for dynamic, editable, high-risk facts. At the \textbf{decision layer}, we separate \emph{fast write routing} from \emph{slow write-back} (the consolidation mechanism). We conceptualize every incoming piece of knowledge through an operational taxonomy of three classes: \textit{non-write}, \textit{write-new}, and \textit{write-update}. Here, \textit{non-write} denotes knowledge already safely answerable from parametric memory, \textit{write-new} denotes knowledge missing from parametric memory, and \textit{write-update} denotes knowledge for which the model produces a non-refusal but incorrect answer, indicating stale or conflicting internal knowledge. 
To manage admission efficiently, the write router bypasses explicit three-way classification and directly optimizes a continuous storage--performance trade-off through a binary admission decision: \emph{write} or \emph{discard}. 
Here, \emph{write} and \emph{discard} are router-level actions used for cost-aware decision-making, while the precise handling of novel versus stale facts is deferred to the subsequent consolidation stage.

%To enable efficient admission control, however, the write router does not explicitly predict these three labels. Instead, it directly optimizes a continuous storage--performance trade-off through a binary routing decision: \emph{write} or \emph{discard}. Here, \emph{discard} is a router-level action used for cost-aware decision-making, rather than an operational knowledge category. The finer distinction between \textit{write-new} and \textit{write-update} is deferred to the subsequent consolidation stage.

A central systems contribution of our framework is that write routing is implemented as a \emph{small-to-large cascade}. Instead of relying on a large model for every routing decision, we use a lightweight model to perform cheap front-end screening and resolve easy cases early, while only uncertain or ambiguous knowledge items are escalated to a stronger model. This design is crucial because write routing is itself a resource-allocation problem: if the router always requires the large model, then routing becomes too expensive to justify. The cascade therefore optimizes not only memory usage, but also the computational cost of the routing policy itself.

To operationalize this formulation, we construct a behavioral labeling pipeline over raw $(\text{knowledge}, \text{query}, \text{answer})$ triples. For each triple, we compare zero-shot answering against answering with explicit memory support. If the model answers correctly zero-shot, the corresponding knowledge is labeled as \textit{non-write}; if the model fails zero-shot but succeeds with memory, it is write-worthy. We further distinguish \textit{write-new} from \textit{write-update} based on whether the zero-shot failure manifests as an explicit refusal or a confident but incorrect answer. These labels allow us to train and evaluate write routers using model behavior rather than abstract semantic judgments.

\begin{figure}[t]
  \centering
  \includegraphics[width=0.85\linewidth]{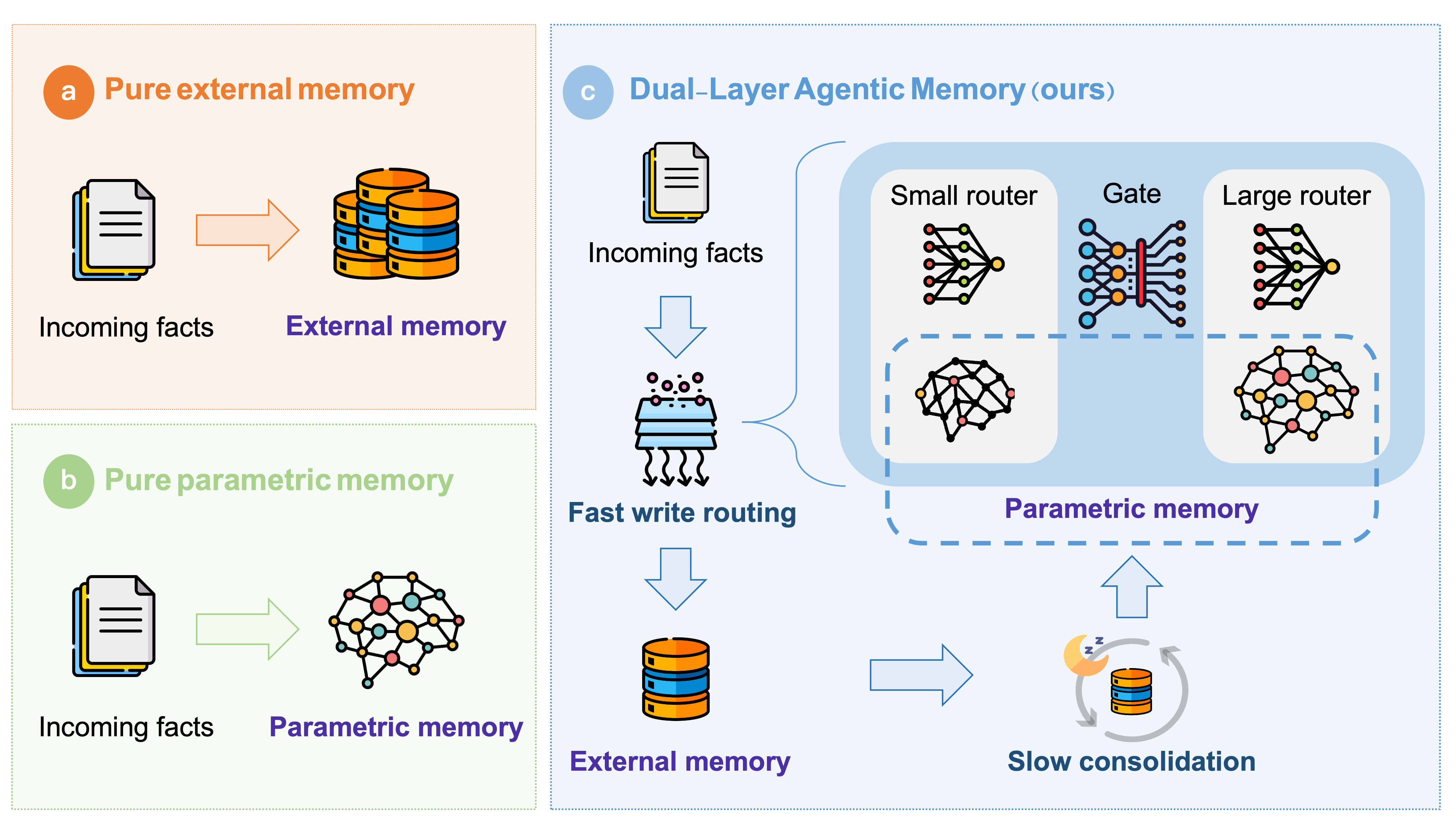}
  \caption{\textbf{Comparison of agent memory paradigms.} 
  \textbf{(a) Pure external memory:} Monotonically appending facts to an external database often leads to memory bloat and degrades retrieval precision. 
  \textbf{(b) Pure parametric memory:} Directly incorporating continuous updates into model weights is computationally expensive and susceptible to catastrophic forgetting. 
  \textbf{(c) Dual-Layer Agentic Memory (ours):} We formulate memory as a knowledge lifecycle. A cost-aware cascade router (\textit{Fast Write Routing}) filters incoming facts, escalating uncertain items to maintain a compact external memory. Retained external memories are periodically internalized via supervised fine-tuning (\textit{Slow Consolidation}). This process updates the model's parametric knowledge, shifting its epistemic boundaries to reduce future reliance on external memory.}
  \label{fig:teaser}
\end{figure}

Finally, we study \emph{write-back} as the slow half of the lifecycle by converting selected \textit{write-new} and \textit{write-update} memories into fine-tuning supervision and testing whether previously retrieval-dependent facts become internally answerable. In this sense, our framework is not only about selective storage, but also about selective internalization: the agent first externalizes only high-value knowledge, and later consolidates stable, validated memories back into parametric memory to reduce future dependence on retrieval. Because the write router operates directly on the model's hidden states and uncertainty profiles, it naturally senses this consolidation; once a fact is internalized, its shifted internal representations prompt the router to discard future encounters of the same knowledge, effectively closing the knowledge lifecycle loop.

Our contributions are summarized as follows:
\begin{itemize}
    \item We formulate agent memory as a \textbf{dual-layer knowledge lifecycle system} that unifies external memory, parametric memory, fast write routing, and slow consolidation.

    \item We introduce an \textbf{operational memory taxonomy} (\textit{non-write}, \textit{write-new}, \textit{write-update}) together with a \textbf{cost-aware small-to-large routing cascade} that frames memory admission as an accuracy--efficiency optimization problem.

    \item We propose a periodic write-back consolidation mechanism that enables \textbf{knowledge internalization} and adaptive \textbf{epistemic shift}, progressively reducing reliance on external memory as parametric knowledge evolves.

    \item We construct an online \textbf{streaming evaluation benchmark} that decouples write-time admission from read-time retrieval, showing that selective externalization and consolidation mitigate memory bloat while closely matching exhaustive-memory performance.

\end{itemize}

\section{Related work}

\label{sec:related}

\paragraph{Memory architectures in LLM-based agents.}
Long-horizon LLM agents typically extend working memory via external RAG \cite{lewis2020retrieval}, with early systems such as Generative Agents and MemGPT introducing long-term memory management \cite{park2023generative, packer2023memgpt}. As memory becomes central to lifelong LLM agents \cite{zheng2026lifelong}, recent systems further improve working-memory compression and memory organization through hierarchical chunking, dynamic linking, and structured memory representations \cite{hu2025hiagent, xu2025mem, zeng2024structural}. Yet these approaches mainly manage active context or organize stored memories, lacking a mechanism to close the knowledge lifecycle loop between external and parametric memory. Consequently, external memory can accumulate redundant facts, leading to memory bloat \cite{salama2025meminsight} and degraded access \cite{liu2024lost} despite structured retrieval mechanisms. In contrast, our Dual-Layer Agentic Memory addresses this by formulating memory as a knowledge lifecycle, where fast write routing and slow consolidation work in tandem to achieve selective externalization followed by selective internalization.

\paragraph{Retrieval-augmented generation and routing.}
To mitigate the latency and noise of exhaustive retrieval, recent frameworks dynamically route memory operations. However, these routing mechanisms operate primarily in the \textit{read} phase, relying on downstream queries to condition retrieval decisions \cite{asai2024self, jeong2024adaptive, labruna2025retrieve, yao2025seakr, wu2025self}. To address the foundational problem of memory admission—deciding what to externalize before storage occurs—we explicitly shift the focus to the \textit{write} phase. Drawing inspiration from cascaded inference, which is widely used to optimize LLM generation \cite{dekoninck2025cascade, ong2025routellm}, we adapt a cost-aware small-to-large cascade to strictly gate memory admission.

\paragraph{Knowledge editing and continual parametric updating.}
Techniques like ROME \cite{meng2022locating} and MEMIT \cite{meng2022mass} update specific model parameters to correct stale facts, while recent methods extend editing to long-form, diverse-format, or parameter-preserving multi-hop settings \cite{jiang2025anyedit, wang2025decoupling, wu2025robust}. However, continuous parametric updates are computationally expensive and highly susceptible to catastrophic forgetting \cite{shi2025continual, wu2024continual}. This motivates using external memory as a fast, editable buffer, with parameter modifications reserved for slower offline consolidation.

\paragraph{Neuroscience-inspired memory and consolidation.}
Our architectural design is motivated by the CLS framework \cite{mcclelland1995there, kumaran2016learning}, modeling the interplay between the fast-learning hippocampus and the slow-learning neocortex. Recent AI systems adopt CLS-inspired consolidation to prevent forgetting through structured memory, reconstructive persistence, or hybrid memory circuits \cite{hsing2026mirror, shi2025hybrid}. Related continual-learning work further shows that sleep-like unsupervised replay can improve retention under limited or imbalanced data \cite{bazhenov2024unsupervised}. Motivated by this consolidation view, we translate the sleep phase into periodic supervised fine-tuning (SFT), physically consolidating transient external memory into the model's parameters.

\section{Dual-Layer Agentic Memory}
\label{sec:method}

We model agent memory as a \emph{dual-layer knowledge lifecycle} problem. Our framework contains two layers. At the \textbf{carrier layer}, the agent maintains two memory substrates: parametric memory $\Theta$, which is stable, compressed, and cheap to access, and external memory $E$, which is explicit, editable, and suited to dynamic or conflict-prone facts. At the \textbf{decision layer}, the system contains a fast \emph{write router}, which decides whether incoming knowledge should enter external memory, and a slow \emph{write-back} mechanism, which selectively internalizes high-value external memories into $\Theta$ through supervised fine-tuning.

\subsection{Operational memory taxonomy}
\label{subsec:taxonomy}

We define write labels based on model behavior under zero-shot versus memory-supported answering. Given a knowledge item $f$ and its associated factual probes, we distinguish three cases:
\begin{itemize}[leftmargin=*, itemsep=2pt]
    \item \textit{non-write}: the agent can already answer the associated query correctly from parametric memory alone, so external memory is unnecessary;
    \item \textit{write-new}: the agent produces a refusal answer in the zero-shot setting, and becomes correct when the relevant knowledge is provided, indicating that the fact is missing from parameters;
    \item \textit{write-update}: the agent produces a non-refusal but incorrect answer in the zero-shot setting, and becomes correct when the relevant knowledge is provided, indicating stale parametric memory.
\end{itemize}

\subsection{Problem setup}
\label{subsec:problem_setup}

Let $f$ denote an incoming knowledge item. At storage time, the router evaluates $f$ to produce a binary admission decision: \emph{discard} (relying solely on parametric memory $\Theta$) or \emph{write} (appending $f$ to external memory $E$). The lifecycle is:
\begin{equation}
    f \xrightarrow{\text{write routing}} 
    \begin{cases}
        \text{discard} \\
        \text{write} \longrightarrow E \xrightarrow{\text{optional write-back}} \Theta'
    \end{cases}
\end{equation}
Subsequently, the items written to $E$ are utilized during the slow write-back stage for parametric consolidation, ultimately updating the model to $\Theta'$.

\subsection{Small-to-large cost-aware write routing}
\label{subsec:write_router}

Write routing is implemented as a \textbf{small-to-large cascade}. Physically, each router consists of a frozen Large Language Model (LLM) backbone---serving as a feature extractor---paired with a trained lightweight Multi-Layer Perceptron (MLP) decision head. A small router $M_{\mathrm{small}}$ (utilizing a smaller LLM backbone) first performs cheap front-end screening and resolves easy cases. Only uncertain or ambiguous examples are escalated to a stronger but more computationally expensive large router $M_{\mathrm{large}}$ (utilizing a larger LLM backbone).

For each fact $f$, $M_{\mathrm{small}}$ predicts whether the fact should be written to external memory ($a_{\mathrm{small}} \in \{W, D\}$). A gating module then estimates whether escalating is worth the additional compute cost. If escalation is triggered, the final decision becomes $a_{\mathrm{large}} \in \{W, D\}$ determined by $M_{\mathrm{large}}$.

\subsubsection{Routing objective}

Let $\lambda_s$ denote the storage penalty and $\lambda_e$ the escalation penalty. Writing a fact is rewarded only when its marginal task-performance gain outweighs the combined penalty $\lambda_s + \lambda_e$. For a given fact, let $\mathrm{EM}(W)$ denote the downstream QA Exact Match (EM) obtained when the fact is written to external memory, and $\mathrm{EM}(D)$ when discarded. We define the write reward as $r(W) = \mathrm{EM}(W) - \mathrm{EM}(D) - \lambda_s$, and the discard reward as $r(D) = 0$.

\subsubsection{Escalation gate}

The small router predicts rewards $\hat r_{\mathrm{small}}(W)$ and $\hat r_{\mathrm{small}}(D)$, and the large router predicts $\hat r_{\mathrm{large}}(W)$ and $\hat r_{\mathrm{large}}(D)$. We define the expected value of stopping at the small router as $V_{\mathrm{small}} = r(a_{\mathrm{small}})$, and the expected value of escalation as $V_{\mathrm{large}} = r(a_{\mathrm{large}}) - \lambda_e$. The gate is trained to predict the gain of escalation: $g^* = V_{\mathrm{large}} - V_{\mathrm{small}}$. At inference time, the example is escalated if and only if the predicted gain is positive ($g_{\mathrm{pred}} > 0$).

\subsubsection{Feature construction and neural parameterization}
The decision heads for $M_{\mathrm{small}}$ and $M_{\mathrm{large}}$ consume semantic and uncertainty features extracted from their respective LLM backbones. Let $\{h_t\}_{t=1}^{L}$ be the final-layer hidden states produced by the LLM for the tokenized fact $f$. We compute a global semantic embedding using mean pooling:
\begin{equation}
    e = \frac{1}{L}\sum_{t=1}^{L} h_t.
\end{equation}
To capture epistemic uncertainty, let $\ell_t = -\log p(x_t \mid x_{<t})$ be the token-level negative log-likelihood. We compute the mean NLL $\bar{\ell} = \frac{1}{L}\sum_{t=1}^{L}\ell_t$, and resample the sequence NLL curve into a fixed-length representation:
\begin{equation}
    c = \mathrm{Interp}_{64}([\ell_1, \ell_2, \dots, \ell_L]).
\end{equation}
The resulting uncertainty vector incorporates the sequence length and the resampled curve:
\begin{equation}
    u =[\bar{\ell}; \log(1+L); c].
\end{equation}

Using these features, the MLP head of the small router ($z_{\mathrm{small}}$), the escalation gate ($z_g$), and the MLP head of the large router ($z_{\mathrm{large}}$) respectively consume:
\begin{align}
    z_{\mathrm{small}} &=[e_{\mathrm{small}}], \\
    z_g &=[e_{\mathrm{small}}; u_{\mathrm{small}}; \hat r_{\mathrm{small}}(W); \hat r_{\mathrm{small}}(D); |\hat r_{\mathrm{small}}(W)-\hat r_{\mathrm{small}}(D)|], \\
    z_{\mathrm{large}} &=[e_{\mathrm{large}}].
\end{align}
All decision heads are parameterized as MLPs (utilizing LayerNorm, GELU activations, and Dropout) trained to regress reward targets using mean squared error, while the LLM backbones remain frozen. At deployment, the policies trace the Pareto frontier across EM--storage--compute trade-offs. \textbf{We defer full stagewise training and deployment protocols to Appendix~\ref{sec:app_training}.}

\subsection{Write-back as slow consolidation}
\label{subsec:writeback}

We introduce \textbf{write-back}, a slow consolidation stage that internalizes selected external memories into model parameters. Let $E = \{f_i\}_{i=1}^{N}$ denote the routed external memory. Each retained fact $f_i$ is associated with a set of factual probes $\mathcal{Q}_i = \{(q_{i1}, a_{i1}), \dots, (q_{im_i}, a_{im_i})\}$. Write-back is applied only to facts labeled \textit{write-new} or \textit{write-update}.

Each selected fact is converted into two complementary forms of supervision: independent question-level examples ($x^{\mathrm{qa}}_{ij} = q_{ij}, y^{\mathrm{qa}}_{ij}=a_{ij}$) and sentence-level targets ($x^{\mathrm{sent}}_{i}, f_i$). We optimize the standard SFT causal language modeling objective over assistant tokens:
\begin{equation}
    \mathcal{L}_{\mathrm{SFT}}(\Theta)
    =
    - \sum_{k=1}^{K} \sum_{t \in \mathcal{A}_k} \log p_{\Theta}(y_{k,t} \mid x_k, y_{k,<t}).
\end{equation}

A fact is removed from external memory only if consolidation is validated, i.e., the updated model answers its probes correctly without retrieval: $\frac{1}{m_i}\sum_{j=1}^{m_i} \mathrm{EM}\large(\mathrm{Ans}_{\Theta'}(q_{ij}), a_{ij}\large) \ge \tau_{\mathrm{flush}}$. Otherwise, it remains in external memory. For successfully consolidated facts, this completes the knowledge lifecycle: initial externalization into $E$, subsequent internalization into $\Theta'$, and final eviction from $E$.

\section{Experiments and results}
\label{sec:experiments}

To evaluate the dual-layer knowledge lifecycle, we construct an online streaming benchmark and compare our framework against various memory admission baselines. 

\subsection{The streaming knowledge lifecycle benchmark}
\label{subsec:benchmark}

Existing memory benchmarks typically present background knowledge and downstream queries simultaneously. This static setup conflates two distinct challenges: deciding \emph{whether} to store an incoming statement, and deciding \emph{what} to retrieve. Because the query is already known when facts are processed, most systems bypass write-time gating entirely—either unconditionally memorizing all inputs or relying exclusively on the downstream query to filter information at read time. To explicitly isolate the admission bottleneck, we cast memory management as an \emph{online streaming task}. By requiring the agent to make immediate write-versus-discard decisions as facts arrive sequentially—strictly prior to encountering any associated queries—we effectively decouple memory admission from retrieval.

\paragraph{Offline dataset and behavioral labeling.}
We derive the fact corpus from the Zero-Shot Relation Extraction (ZsRE) benchmark \cite{levy2017zeroshot}. To establish targets for memory admission, we probe a frozen \texttt{qwen3-8B} baseline on $6{,}040{,}560$ fact-related queries. By comparing zero-shot answers against memory-supported answers, we quantify the model's parametric blind spots. Based on these behavioral signals, we map each fact into an operational taxonomy: \textit{non-write}, \textit{write-new}, or \textit{write-update} (see Appendix~\ref{sec:app_dataset} for mathematical formulations and label distributions).

% \paragraph{Online benchmark environment and trajectory generation.}
% The online benchmark assesses continuous memory management across $E=300$ independent test-split episodes. To simulate realistic agent scenarios, each episode spans $T=500$ temporally interleaved conversational turns where knowledge injections and downstream queries naturally co-occur. Within each episode, $K=100$ candidate facts are streamed to the agent. A typical interaction trajectory is formally structured is presented in Figure~\ref{fig:streaming_format} in Appendix~\ref{sec:app_dataset}. This explicitly interleaved format simulates real-world agent scenarios characterized by continuous knowledge injection and multi-turn QA interactions. Under this streaming setting, the system must dynamically decide whether each incoming fact warrants externalization strictly at write-time. The injection sequence is synthesized to satisfy causal admissibility (i.e., a query never precedes its injected fact) while strictly preserving the natural proportion of the offline taxonomy. This inherently includes \textit{non-write} distractors to evaluate admission precision. The formal episode synthesis algorithm is deferred to Appendix~\ref{sec:app_dataset}.
\paragraph{Online benchmark environment and trajectory generation.}
We evaluate continuous memory management over $E=300$ streaming episodes from the test split. Each episode contains $T=500$ temporally interleaved conversational turns with $K=100$ candidate fact injections and downstream QA interactions. This streaming setup simulates realistic agent environments where knowledge injection and reasoning co-occur, requiring the system to decide at write-time whether each incoming fact should be externalized. The injected facts preserve the natural distribution of the offline taxonomy, including \textit{non-write} distractors, while satisfying causal admissibility (i.e., queries never precede supporting facts). A typical interaction trajectory is shown in Figure~\ref{fig:streaming_format}, and the full synthesis procedure is provided in Appendix~\ref{sec:app_dataset}.
% \begin{itemize}[label={}, leftmargin=1.5em, itemsep=0pt, parsep=2pt]
%     \item \textbf{User:} $[\text{Fact } f_{i}] \dots[\text{Fact } f_{j}] + [\text{Query } q_{1}]$
%     \item \textbf{Agent:} $[\text{Answer } a_{1}]$
%     \item \textbf{User:} $[\text{Query } q_{2}]$
%     \item \textbf{Agent:} $[\text{Answer } a_{2}]$
%     \item \dots
%     \item \textbf{User:} $[\text{Fact } f_{k}] +[\text{Query } q_{3}]$
%     \item \textbf{Agent:} $[\text{Answer } a_{3}]$
% \end{itemize}

\paragraph{Metrics.}
We evaluate downstream QA quality using Exact Match (EM), token-level $F_1$, and refusal rate. Storage decision quality is scored at the knowledge-injection level against the fact-level labels via storage ratio together with Store Precision, Store Recall, and Store $F_1$ (see Appendix~\ref{sec:app_dataset} for specific details on memory admission targets). Finally, system efficiency is reported as routing compute cost.

\subsection{Baselines and compared methods}
\label{subsec:baselines}

We compare against \textbf{Static Policies} (\texttt{No Store}, \texttt{Full Store}, \texttt{Random Store}), \textbf{Heuristic Policies} (\texttt{Heuristic Store}, \texttt{PPL-Conditional Store}), and \textbf{Supervised Policies} (\texttt{Logistic Regression}, \texttt{MLP Classifier}). For our cascaded methods, we explicitly instantiate $M_{\mathrm{small}}$ using \texttt{qwen3-1.7B} as the backbone, and $M_{\mathrm{large}}$ using \texttt{qwen3-8B} \cite{yang2025qwen3}. We evaluate \textbf{\texttt{Write Router}}, which operates on features extracted from the frozen base models, and \textbf{\texttt{Write Router}$_{\text{SFT}}$}, where both the 1.7B and 8B backbones have undergone the write-back consolidation phase to maintain aligned knowledge distributions. \textbf{Detailed implementation and thresholding strategies for all heuristic and supervised baselines are provided in Appendix~\ref{sec:app_baselines}.} All experiments are conducted on a single compute node equipped with 4 NVIDIA H20 GPUs.

\subsection{Main results: end-to-end online benchmark}
\label{subsec:main_results}

We evaluate all methods on the online streaming benchmark. Table~\ref{tab:main_benchmark} reports downstream QA quality together with storage and routing compute efficiency. The overarching question is whether our dual-layer framework can approach the performance of exhaustive storage while simultaneously avoiding external memory bloat and exorbitant routing costs.

\begin{table}[h]
  \caption{End-to-end results on the dynamic online benchmark. Our cascaded \texttt{Write Router} improves the storage--performance frontier by reducing both routing compute and external memory usage: it escalates only $\sim$40--49\% of inputs to the 8B model, prunes up to 68\% of redundant memory, and still retains over 98.2\% of \texttt{Full Store} EM.}
  \label{tab:main_benchmark}
  \centering % 确保表格居中
  \footnotesize % 保持适中字号
  \setlength{\tabcolsep}{3pt} % 压缩列间距以适应页面宽度
  \begin{tabular}{@{} l ccc cccc l @{}} % 首尾的 @{} 可以去掉表格多余的空白边距，保证完美居中不溢出
    \toprule
    \textbf{Policy} & \makecell{\textbf{QA}\\\textbf{EM}} & \makecell{\textbf{Tok.}\\\textbf{$F_1$}} & \makecell{\textbf{Ref.}\\\textbf{Rate}} & \makecell{\textbf{Store}\\\textbf{($\downarrow$)}} & \makecell{\textbf{Store}\\\textbf{Prec.}} & \makecell{\textbf{Store}\\\textbf{Recall}} & \makecell{\textbf{Store}\\\textbf{$F_1$}} & 
    \textbf{Routing Cost$^\dagger$ ($\downarrow$)} \\
    %\makecell{\textbf{Routing Cost$^\dagger$}\\\textbf{($\downarrow$)}} \\
    \midrule
    \multicolumn{9}{c}{\textit{Base Model Framework (qwen3-8B)}} \\
    \midrule
    \texttt{No Store}                 & 38.02\% & 0.4039 & 2.8\% & \textbf{0.0\%}  & --      & 0.0\%   & --      & \textbf{0} \\
    \texttt{Random Store} $(p=0.5)$   & 62.95\% & 0.6517 & 1.6\% & \underline{50.0\%}  & 76.63\% & 50.06\% & 60.56\% & \textbf{0} \\
    \texttt{Random Store} $(p=0.8)$   & 78.07\% & 0.8020 & 0.7\% & 80.0\%  & 76.60\% & 80.07\% & 78.30\% & \textbf{0} \\
    \texttt{Full Store}               & \textbf{87.79\%} & \textbf{0.8980} & \textbf{0.1\%} & 100.0\% & 76.54\% & \textbf{100.0\%} & 86.71\% & \textbf{0} \\
    \texttt{Heuristic Store}          & 73.56\% & 0.7605 & 0.6\% & 59.08\% & 86.06\% & 66.43\% & 74.98\% & $2\times \text{Gen}_{8\text{B}}$ \\
    \texttt{PPL-Cond. Store}          & 78.95\% & 0.8144 & 0.5\% & 71.74\% & 83.58\% & 78.35\% & 80.88\% & $1\times \text{Fwd}_{8\text{B}}$ \\
    \texttt{Logistic Regression}      & 83.84\% & 0.8611 & \underline{0.2\%} & 68.50\% & 95.68\% & 85.64\% & \underline{90.38\%} & $1\times \text{Fwd}_{8\text{B}}$ \\
    \texttt{MLP Classifier}           & 83.18\% & 0.8548 & 0.3\% & 65.36\% & \textbf{96.87\%} & 82.72\% & 89.24\% & $1\times \text{Fwd}_{8\text{B}}$ \\
    \makecell[l]{\texttt{Write Router} \\ $(\lambda_s=0.2,\, \lambda_e=0.008)$} & 83.97\% & 0.8622 & 0.3\% & 64.81\% & \underline{95.76\%} & 81.09\% & 87.82\% & $\text{Fwd}_{1.7\text{B}} \!+\! {49.0\%}\, \text{Fwd}_{8\text{B}}$ \\
    \makecell[l]{\texttt{Write Router} \\ $(\lambda_s=0.08,\, \lambda_e=0.005)$}& \underline{86.35\%} & \underline{0.8846} & \underline{0.2\%} & 77.00\% & 91.06\% & \underline{91.61\%} & \textbf{91.33\%} & \underline{$\text{Fwd}_{1.7\text{B}} \!+\! {39.7\%}\, \text{Fwd}_{8\text{B}}$} \\
    \midrule
    \multicolumn{9}{c}{\textit{SFT Model Framework (Knowledge-adapted qwen3-8B)}} \\
    \midrule
    \texttt{No Store}                 & 69.16\% & 0.7154 & 0.0\% & \textbf{0.0\%}   & --      & 0.0\%   & --      & \textbf{0} \\
    \texttt{Full Store}               & \textbf{92.29\%} & \textbf{0.9303} & 0.0\% & 100.0\% & 76.54\% & \textbf{100.0\%} & \textbf{86.71\%} & \textbf{0} \\
    \makecell[l]{\texttt{Write Router}$_{\text{SFT}}$ \\ $(\lambda_s=0.2,\, \lambda_e=0.008)$} & 89.77\% & 0.9169 & {0.0\%} & \underline{32.08\%} & \textbf{97.85\%} & 41.01\%$^*$ & 57.80\% & $\text{Fwd}_{1.7\text{B}} \!+\! {45.8\%}\, \text{Fwd}_{8\text{B}}$ \\
    \makecell[l]{\texttt{Write Router}$_{\text{SFT}}$ \\ $(\lambda_s=0.08,\, \lambda_e=0.005)$}& \underline{90.71\%} & \underline{0.9253} & {0.0\%} & 47.85\% & \underline{93.07\%} & \underline{58.19\%}$^*$ & \underline{71.61\%} & \underline{$\text{Fwd}_{1.7\text{B}} \!+\! {43.3\%}\, \text{Fwd}_{8\text{B}}$}\\
    \bottomrule
\end{tabular}
  \vspace{2pt}
  \raggedright
  % 脚注部分
  % $^\dagger$Routing cost compares inference overhead. ``$\text{Gen}_{8\text{B}}$'' denotes an autoregressive generation call (up to 50 steps per sequence) with the 8B model. ``$\text{Fwd}_{8\text{B}}$'' and ``$\text{Fwd}_{1.7\text{B}}$'' denote a single parallelized forward pass of the 8B and 1.7B models, respectively. Our cascaded \texttt{Write Router} processes all inputs with the 1.7B model and escalates only a fraction (shown as percentages) to the 8B model, avoiding the strict $1\times \text{Fwd}_{8\text{B}}$ cost of standard feature extractors.
  
  % $^*$The Store Recalls from $\texttt{Write Router}_{\text{SFT}}$ are evaluated against static base-model targets.
  $^\dagger$Routing cost measures inference overhead. $\text{Gen}_{8\text{B}}$ denotes autoregressive generation with the 8B model, while $\text{Fwd}_{8\text{B}}$ and $\text{Fwd}_{1.7\text{B}}$ denote a single forward pass. The cascaded \texttt{Write Router} filters all inputs with the 1.7B model and escalates only a subset to the 8B model.

$^*$Store Recall for $\texttt{Write Router}_{\text{SFT}}$ is evaluated against static base-model targets.
\end{table}

\paragraph{Limitations of static and heuristic policies.}
Under the base model framework, relying purely on parametric memory yields poor performance, whereas exhaustive retention provides a theoretical upper bound. As shown in Table~\ref{tab:main_benchmark}, while \textbf{Static Policies} like \texttt{Random Store} can partially recover accuracy, they are highly memory-inefficient. \textbf{Heuristic Policies} (\texttt{Heuristic Store} and \texttt{PPL-Conditional Store}) improve the storage--performance frontier but remain suboptimal compared to \textbf{Supervised Policies} (\texttt{Logistic Regression} and \texttt{MLP Classifier}). This indicates that optimal write decisions require explicit training on model-internal features.

\paragraph{Dual efficiency via cascade routing.} 
Beyond storage footprint, the inference overhead of routing is a critical bottleneck. Explicit probing (\texttt{Heuristic Store}) and full-model feature extraction (\textbf{Supervised Policies}) incur substantial computational costs. Our \textbf{\texttt{Write Router}} addresses this through a small-to-large cascade. By deploying \texttt{qwen3-1.7B} as an efficient proxy, it escalates only 39.7\% to 49.0\% of uncertain queries to the 8B model, reducing the expected inference cost by roughly 30\% to 39\%. Furthermore, it establishes a strictly superior storage--performance frontier for the base model. It decisively outperforms not only static and heuristic baselines but also full-8B \textbf{Supervised Policies} at a fraction of the compute cost. Ultimately, the \texttt{Write Router} tightly tracks the theoretical optimum, retaining over 98.3\% of the exhaustive retention upper bound (\texttt{Full Store}) while significantly reducing both memory bloat and routing compute overhead.

\paragraph{Knowledge internalization and epistemic shift.}
Table~\ref{tab:main_benchmark} (bottom) evaluates \textbf{\texttt{Write Router}$_{\text{SFT}}$}. Periodic SFT facilitates \textbf{knowledge internalization}, significantly improving baseline QA performance and inducing an \textbf{epistemic shift}. Sensing this updated parametric capacity, the router correctly discards newly internalized facts---a behavior that mathematically manifests as an apparent drop in Store Recall when evaluated against static base-model targets. Together, this adaptive discarding and the precise targeting of remaining blind spots (97.85\% Store Precision) allow the router to drastically reduce the storage footprint to just 32.08\%. Despite this minimal external memory reliance, the agent achieves 90.71\% QA EM, retaining over 98.2\% of the \texttt{Full Store} upper bound.

\subsection{Cascade gating analysis and architectures}
\label{subsec:ablations_summary}

% To determine the optimal cascade parameterization, we conducted extensive ablations on the held-out test split of our offline routing dataset. We found that combining semantic features with epistemic uncertainty using mean pooling and Mean Squared Error (MSE) optimization yields the strongest Pareto frontier (detailed in Appendix~\ref{sec:app_ablations}). Crucially, we evaluate whether the thresholded small-to-large cascade can recover the performance of the large router without incurring its full computational cost. Figure~\ref{fig:main_test_frontier} visualizes this trade-off against the constituent small and large router baselines.

% As shown in Figure~\ref{fig:main_test_frontier}, the thresholded cascade consistently closes the performance gap between the small and large routers across all relevant storage capacities. Even under strict escalation budgets, the cascade frontier nearly perfectly tracks the theoretical upper bound set by the large router, incurring only negligible EM degradation while offering substantial accuracy improvements over the standalone small router (see Appendix~\ref{sec:appendix_cascade_numbers} for detailed numerical breakdowns). This empirically validates our core design: escalating only a moderate fraction of uncertain inputs is sufficient to preserve near-optimal routing quality, thereby decisively cutting inference costs for memory admission.
\begin{wrapfigure}{r}{0.64\textwidth}
% \begin{figure}[h]
  \centering
  \includegraphics[width=0.70\linewidth]{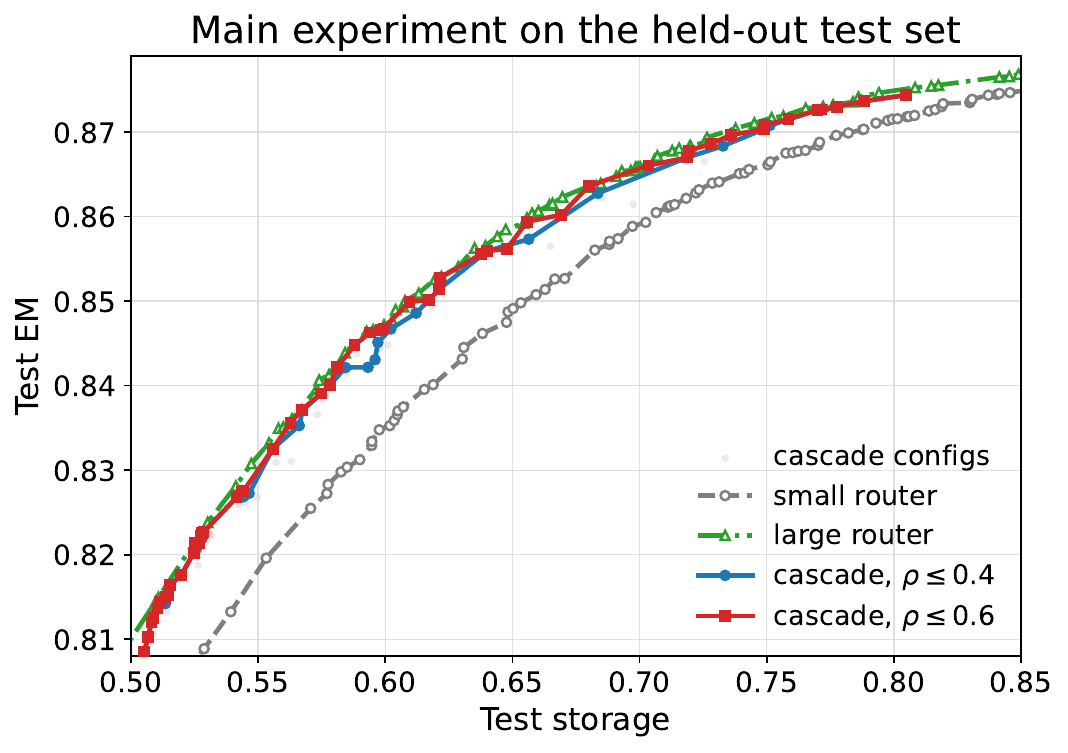}
  % \caption{Held-out test-set EM--storage Pareto frontiers for the main experimental configuration, \texttt{cascade\_gate\_semantic\_uncertainty}, under escalation thresholds $\rho \in \{0.4, 0.6\}$. The thresholded cascade strongly tracks the upper-bound large router while dramatically reducing compute overhead.}
  \caption{EM--storage Pareto frontiers for the main cascade configuration under escalation thresholds $\rho \in \{0.4, 0.6\}$.}
  \label{fig:main_test_frontier}
% \end{figure}
\end{wrapfigure}

To determine the optimal cascade parameterization, we conduct ablations on the held-out routing test split. Semantic features combined with epistemic uncertainty, mean pooling, and MSE optimization yield the strongest Pareto frontier (detailed in Appendix~\ref{sec:app_ablations}). Figure~\ref{fig:main_test_frontier} evaluates whether the thresholded small-to-large cascade can recover large-router performance at lower computational cost.

As shown in Figure~\ref{fig:main_test_frontier}, the cascade consistently closes the small--large router gap across storage budgets. Even under strict escalation limits, it closely tracks the large-router upper bound, with negligible EM degradation and substantial gains over the standalone small router (see Appendix~\ref{sec:appendix_cascade_numbers}). This validates our design: escalating only uncertain inputs preserves near-optimal routing quality while reducing memory-admission inference cost.

\subsection{Write-back and consolidation analysis}
\label{subsec:writeback_results}

The final lifecycle stage, write-back, converts selected external memories into SFT supervision. The objective is \emph{selective internalization}—embedding factual updates into parametric memory ($\Theta'$) to reduce subsequent retrieval dependence. As shown previously, the write-back phase alters the agent's baseline QA performance and reduces the required storage ratio from 64.81\% to 32.08\%.

\begin{table}[ht]
  \caption{Post-write-back label transition matrix. While many facts become \textit{non-write}, 1,752 previously stable facts degrade, indicating parametric interference.}
  \label{tab:transition_matrix}
  \centering
  \small
  \begin{tabular}{lcccr}
    \toprule
    \textbf{Base \textbackslash{} SFT} & \textbf{non-write} & \textbf{write-new} & \textbf{write-update} & \textbf{Total (Base)} \\
    \midrule
    \textbf{non-write}    & 5,347  & 0 & 1,752 & 7,099 \\
    \textbf{write-new}    & 5,785  & 0 & 7,939 & 13,724 \\
    \textbf{write-update} & 4,898  & 0 & 4,482 & 9,380 \\
    \midrule
    \textbf{Total (SFT)}  & 16,030 & 0 & 14,173& \textbf{30,203} \\
    \bottomrule
  \end{tabular}
\end{table}

To analyze this epistemic shift at the fact level, we track the behavioral labels of 30,203 held-out knowledge items before and after consolidation in Table~\ref{tab:transition_matrix}. The SFT stage alters the labels of 67.46\% of the knowledge base, revealing three distinct phenomena:

\begin{itemize}[leftmargin=*, itemsep=2pt, parsep=0pt]
    \item \textbf{Parametric Internalization:} A total of 10,683 retrieval-dependent facts (5,785 \textit{write-new} and 4,898 \textit{write-update}) transition to \textit{non-write}. This indicates that the model can now answer these queries using parametric memory alone, effectively reducing reliance on external memory.
    \item \textbf{Partial Absorption:} 7,939 \textit{write-new} facts shift to \textit{write-update}. Here, the model moves from explicit zero-shot refusals to generating incorrect answers. This suggests that the model acquires partial semantic familiarity during SFT but still relies on external memory support for exact factual correction.
    \item \textbf{Parametric Interference:} We observe that 1,752 previously stable \textit{non-write} facts degrade into \textit{write-update}. This reflects the catastrophic forgetting common in continuous parametric memory updating. This interference provides empirical motivation for the dual-layer architecture: because slow consolidation can corrupt existing knowledge, maintaining an external memory layer is necessary to override newly induced parametric errors.
\end{itemize}

\section{Discussion \& conclusion}
\label{sec:discussion_conclusion}

In this work, we formalize LLM agent memory not as a passive, monotonically growing repository, but as a dynamic \emph{knowledge lifecycle} problem. Drawing inspiration from complementary learning systems, our dual-layer architecture integrates fast, selective write routing with slow parametric consolidation. Our empirical evaluations yield three primary conclusions. First, \textbf{cost-aware selective externalization} via a small-to-large cascaded router efficiently filters redundant facts, preserving downstream accuracy near the theoretical upper bound while drastically reducing both external memory bloat and routing compute overhead. Second, \textbf{write-back enables internalization}, leveraging the external memory as a temporary buffer that allows the agent to internalize retrieval-dependent facts through periodic SFT. Because the router is conditioned on the model's internal states, it adaptively recognizes this shifted parametric capacity and further suppresses the external memory ratio. Finally, \textbf{parametric interference motivates the dual-layer design}: our fact-level transition analysis reveals that while continuous parametric updates internalize vast knowledge, they inherently disrupt a subset of previously stable facts. This confirms the necessity of maintaining a fast-routing external memory layer: it acts as an essential buffer to dynamically correct the inevitable epistemic regressions caused by periodic parametric consolidation.

\paragraph{Limitations and future work.}
While our framework provides a principled memory lifecycle, several limitations motivate future work. First, the offline SFT write-back introduces internalization latency and high compute costs; exploring continuous, parameter-efficient adaptation (e.g., online LoRA) could tighten this loop. Second, treating the external buffer as a uniform dataset during consolidation lacks a principled mechanism to resolve temporally conflicting updates accumulated between cycles. Finally, while the external layer effectively buffers parametric interference, actively mitigating this catastrophic forgetting during write-back—such as via episodic replay or weight regularization—remains a critical challenge for lifelong agent learning.

\begin{ack}
%[AUTHOR NOTE: Acknowledgments go here. Do NOT include this section in the anonymized submission.]
\end{ack}

\bibliographystyle{unsrtnat}
\bibliography{references}

%%%%%%%%%%%%%%%%%%%%%%%%%%%%%%%%%%%%%%%%%%%%%%%%%%%%%%%%%%%%

\newpage
\appendix

\section{Extended dataset and benchmark details}
\label{sec:app_dataset}

This section details the mathematical formulations, empirical distributions, and algorithmic synthesis of the streaming benchmark introduced in Section~\ref{subsec:benchmark}.

\subsection{Offline routing dataset and behavioral supervision}
\label{subsubsec:corpus}

We build on the Zero-Shot Relation Extraction (ZsRE) benchmark~\cite{levy2017zeroshot}, which pairs each factual \emph{(subject, relation, object)} triple with crowd-sourced paraphrases of the underlying relation. After enforcing a strict ``exactly $m{=}30$ paraphrased probes per fact'' filter, we retain $N{=}201{,}352$ unique fact instances $f_i$, each accompanied by a set of factual probes $\mathcal{Q}_i = \{(q_{ij}, a_{ij})\}_{j=1}^{m}$. The queries ($q_{ij}$) within these $m$ diverse probes evaluate the model's robustness to linguistic variations of the same underlying knowledge, yielding a total of $Nm{=}6{,}040{,}560$ distinct probes.

To establish behavioral labels that serve as evaluation targets for memory admission, analytical markers for tracking parametric memory shifts, and supervision for classifier baselines, we evaluate a frozen baseline LLM (\texttt{qwen3-8B}, greedy decoding) twice per probe: once \emph{zero-shot} (using only the query $q_{ij}$) and once \emph{memory-supported} (where the corresponding fact $f_i$ is prepended to $q_{ij}$). Comparing the two generated responses against the gold answer $a_{ij}$ partitions the probes into four operational classes: \textit{internal-answerable} (correct zero-shot), \textit{missing-knowledge} (refused zero-shot but correct with the fact; corresponding to $m_{\text{new}}$), \textit{stale-knowledge} (incorrect zero-shot but correct with the fact; corresponding to $m_{\text{update}}$), and \textit{unsolved} (incorrect even when the fact $f_i$ is provided). The substantial proportion of \textit{stale-knowledge} cases ($19.4\%$ of probes) is particularly informative: it identifies a large slice of parametric knowledge that is not merely missing but \emph{stale}, which an effective memory must \emph{override} rather than simply supplement.

To aggregate these probe-level signals into a fact-level taxonomy, we compute the overall fraction of probes requiring external memory support ($\rho_i$) and the fraction of these memory-dependent probes indicating stale parametric memory ($\gamma_i$):
\begin{equation}
    \rho_i = \frac{m_{\text{new}} + m_{\text{update}}}{m}, \quad \gamma_i = \frac{m_{\text{update}}}{m_{\text{new}} + m_{\text{update}}}
\end{equation}
Based on these metric ratios, the fact-level write label $\ell_i$ is strictly determined as follows: \textit{non-write} if $\rho_i < 0.01$; \textit{write-update} if $\rho_i \ge 0.01$ and $\gamma_i \ge 0.5$; and \textit{write-new} otherwise ($\rho_i \ge 0.01$ and $\gamma_i < 0.5$). This deterministic assignment provides the ground-truth labels for training and evaluating the memory admission policies. The empirical class proportions are \textit{non-write}{:}\textit{write-new}{:}\textit{write-update} = 23.5{:}\,45.1{:}\,31.4\,(\%). The inclusion of a non-trivial \textit{non-write} fraction is essential to penalize indiscriminate memory admission.

\subsection{Streaming evaluation protocol and episode synthesis}
\label{subsubsec:synthesis}

We instantiate $E{=}300$ test-split \emph{episodes}, each modeled as a finite-horizon environment of $T{=}500$ turns over a private fact pool of size $K{=}100$. Facts are drawn without replacement from the test split to ensure no fact appears in more than one episode. A typical interaction trajectory is formally structured as follows:
\begin{figure}[h]
\centering
\includegraphics[width=0.5\linewidth]{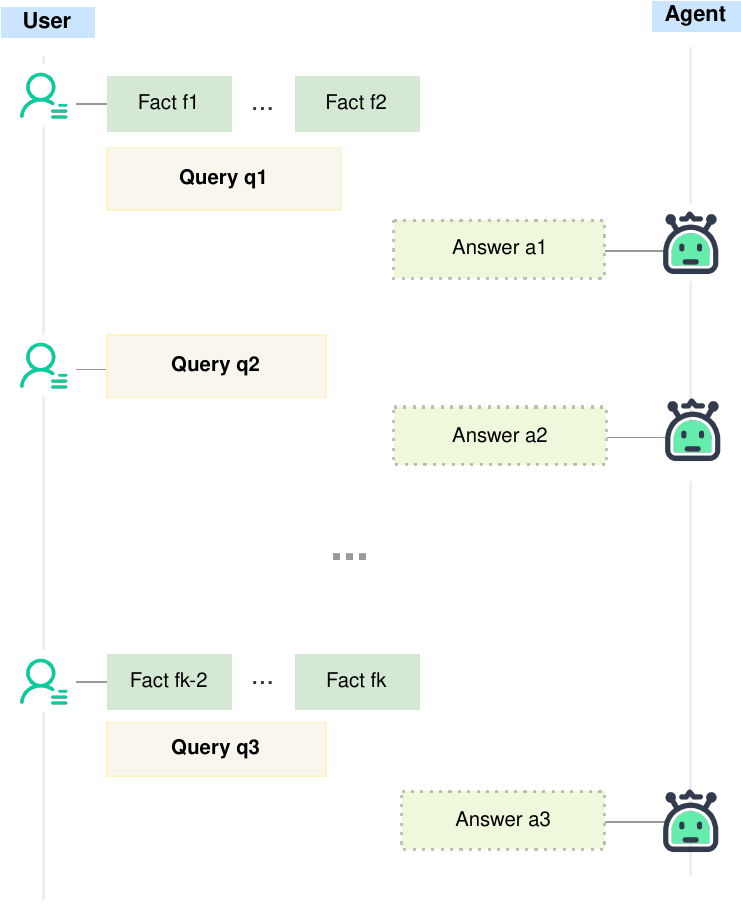}
\caption{Illustration of the streaming interaction format used in our benchmark.}
\label{fig:streaming_format}
\end{figure}

Episodes are synthesized automatically by a scheduler that samples from the fact pool subject to four critical properties:
\begin{description}\setlength{\itemsep}{2pt}
\item[(P1) Coverage.] Every sampled fact is injected exactly once and probed at least once.
\item[(P2) Causality.] No query $q_{kj}$ testing $f_k$ may appear before the turn at which $f_k$ is injected.
\item[(P3) Liveness.] Turn $0$ admits at least one knowledge injection, ensuring the first query is causally answerable.
\item[(P4) Distractor exposure.] The empirical class mix matches the underlying test-split distribution, exposing the agent to \textit{non-write} distractors at their natural base rate.
\end{description}

\paragraph{Algorithm design and guarantees.}
To satisfy these temporal and causal constraints efficiently, we employ a three-stage deterministic generation pipeline (Algorithm~\ref{alg:synth}) driven by a seeded random number generator. 

First, facts are distributed across the timeline using a stratified sampling approach (\textbf{Stage 1}). Instead of randomly assigning injection times, the timeline is partitioned into buckets, guaranteeing that every fact is injected early enough to leave sufficient room for subsequent probes. A post-hoc check ensures at least one fact is injected at turn 0 to satisfy liveness (P3). 

Second, the algorithm allocates a precise probe budget to each fact (\textbf{Stage 2}). It uses a forward greedy pass to dynamically assign probe counts, ensuring the total number of queries perfectly matches the episode length ($T$). Because of the stratified injection in Stage 1, the algorithm inherently avoids bottlenecks where too many facts are injected too late to be probed.

Finally, probes are sequentially streamed to the agent (\textbf{Stage 3}). We maintain an active pool of valid queries ($\mathcal{P}$). A query only enters this active pool at or after the exact turn its corresponding fact is injected. By strictly sampling queries from $\mathcal{P}$, the algorithm naturally enforces causal admissibility (P2). 

This pipeline runs in $O(T \log T)$ time and provides a highly robust, bit-exactly reproducible mechanism. By relying on structural scheduling rather than complex constraint solvers, it seamlessly simulates continuous knowledge evolution while strictly respecting all evaluation requirements.

\begin{algorithm}[ht]
\caption{Episode synthesis $(K, T, m, \mathrm{rng})$.}
\label{alg:synth}
\begin{algorithmic}[1]
\Statex \textbf{Stage 1: Stratified injection schedule}
\For{$k = 1, \dots, K$}
  \State $l_k \!\gets\! \lfloor (k{-}1)T/K\rfloor$,\ \ $r_k \!\gets\! \min\{T{-}1,\lfloor kT/K\rfloor{-}1\}$
  \State $\iota(k) \!\sim\! \mathcal{U}\{l_k,\dots,r_k\}$ \Comment{Assign fact injection turns $\iota(k)$}
\EndFor
\If{$0 \notin \iota([K])$}\ \ $\iota(\arg\min_k \iota(k)) \gets 0$ \EndIf
\Statex \textbf{Stage 2: Probe budget allocation}
\State $\sigma \gets$ permutation of $[K]$ in ascending $\iota$;\ \ $n \gets \mathbf{1}_K$;\ \ $c \gets K$
\For{$r = 1, \dots, K$}
  \State $t^\star \gets \iota(\sigma(r{+}1))$ if $r<K$ else $T$;\ \ $s \gets t^\star - c$
  \State \textbf{advance} $n_{\sigma(r)}, n_{\sigma(r{-}1)}, \dots$ in turn (up to cap $m$) until $s$ is absorbed, updating $c$
\EndFor
\State Distribute the residual $T - \sum_k n_k$ uniformly at random subject to $n_k \le m$
\Statex \textbf{Stage 3: Causal probe streaming}
\For{$k = 1, \dots, K$}\ \ $P_k \gets$ uniform sample of $n_k$ probes from $\mathcal{Q}_k$ \EndFor
\State $\mathcal{P} \gets \emptyset$
\For{$t = 0, \dots, T{-}1$}
  \State $\mathcal{P} \gets \mathcal{P} \cup \{(k,p): p \in P_k,\, \iota(k){=}t\}$ \Comment{Add probe to pool only after injection}
  \State $\pi(t) \gets$ uniform sample from $\mathcal{P}$;\ \ $\mathcal{P} \gets \mathcal{P} \setminus \{\pi(t)\}$
\EndFor
\State \Return $(\iota, n, \pi)$
\end{algorithmic}
\end{algorithm}

\subsection{Evaluation metrics}
We report three families of metrics. Downstream QA quality is scored per turn by evaluating the agent's generated response to the current query against the corresponding ground-truth answer using Exact Match (EM), token-level $F_1$, and refusal rate. Storage decision quality is scored at the knowledge-injection level against the fact-level labels $\ell_i$ via storage ratio together with Store Precision, Store Recall, and Store $F_1$. Here, behavioral labels \textit{write-new} and \textit{write-update} are merged and treated as positive targets for memory admission, while \textit{non-write} acts as the negative target. Finally, system efficiency is reported as routing compute cost (inference overhead per decision).

\section{Implementation of baselines and compared methods}
\label{sec:app_baselines}
Due to space constraints in the main text, we detail the implementation of heuristic and supervised baselines here:
\begin{itemize}[leftmargin=*]
    \item \textbf{\texttt{Heuristic Store}}: Explicitly probes the model before making a write decision. Given an incoming fact, it generates a factual question, answers that question, and uses a high/low confidence judgment to decide whether the fact should be stored.
    \item \textbf{\texttt{PPL-Conditional Store}}: Thresholds the model's per-token cross-entropy on the input fact, using linguistic surprisal as a proxy for prior familiarity. The threshold is selected via a validation split.
    \item \textbf{\texttt{Logistic Regression} \& \texttt{MLP Classifier}}: Trained on automatically constructed binary \textit{write} vs. \textit{non-write} labels. The \texttt{MLP Classifier} shares the exact same architecture as our large router $M_{\mathrm{large}}$ and uses the same semantic/uncertainty features, but is optimized with standard binary cross-entropy (BCE) instead of our reward-based objective.
\end{itemize}

\section{Training and deployment protocols}
\label{sec:app_training}
Training proceeds stagewise. We first train the small router and the large router to regress reward targets using mean squared error. After freezing both routers, we construct gate supervision from their greedy actions and train the gate to predict escalation gain. In deployment, routing policies are not chosen by maximizing QA Exact Match (EM) alone, since that would trivially collapse toward writing everything. Instead, we sweep $(\lambda_s, \lambda_e)$ on the validation split, tracing the Pareto frontier in the EM-storage-compute space to select the optimal deployment trade-off.

\section{Extended router architectural ablations}
\label{sec:app_ablations}

We analyzed the write router in the controlled offline setting to isolate the contributions of feature design, pooling strategy, loss function, and cascade gating design. 

\subsection{Feature ablation}
We performed a $3 \times 3$ ablation study comparing feature sources from $M_{\mathrm{small}}$, $M_{\mathrm{large}}$, and their concatenation against semantic, uncertainty, and combined features. 

\begin{figure}[ht]
  \centering
  \includegraphics[width=0.9\linewidth]{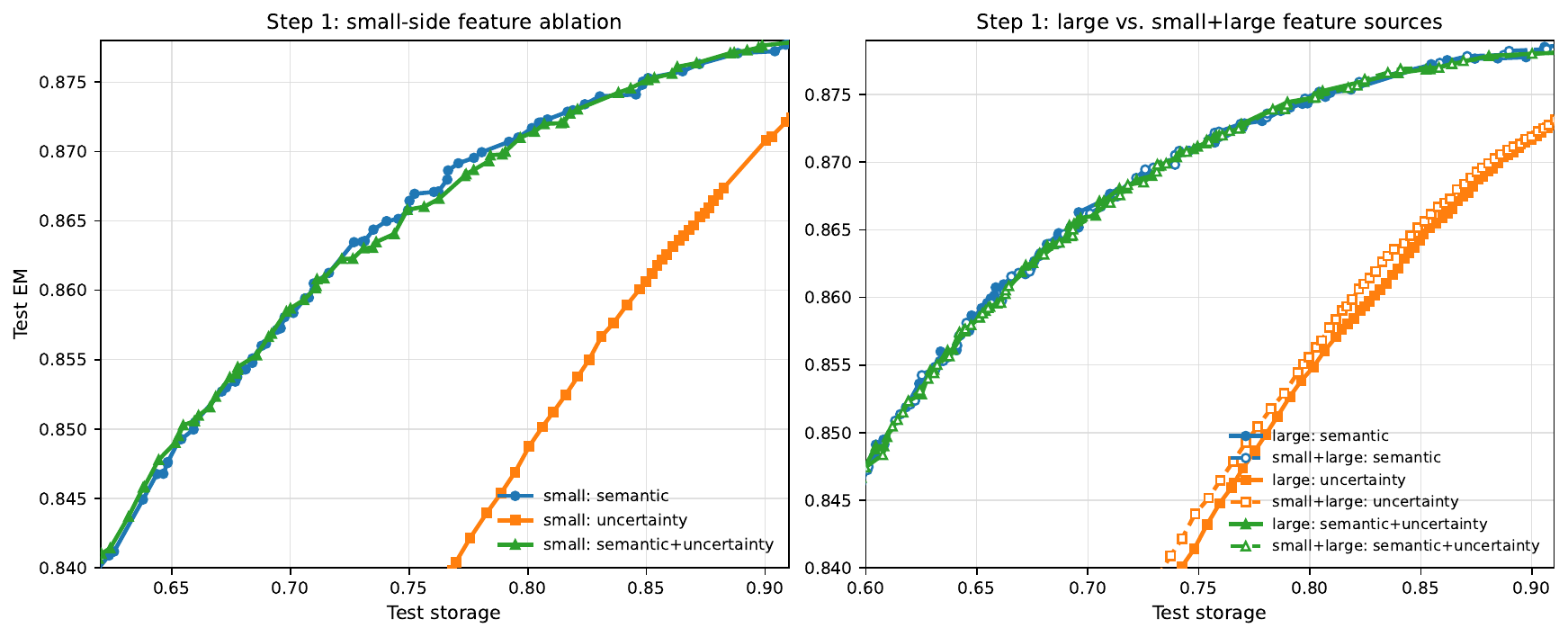}
  \caption{Step 1 single-layer feature ablation on the held-out test set.}
  \label{fig:ablation_step1}
\end{figure}

As shown in Figure~\ref{fig:ablation_step1}, uncertainty-only inputs are clearly weaker than semantic inputs on the small side: to reach test EM $0.86$, \texttt{single\_small\_uncertainty} requires storage $0.847$, whereas \texttt{single\_small\_semantic} requires only $0.710$. On the large router, concatenating small-side features offers almost no benefit: at storage $\le 0.8$, \texttt{single\_large\_semantic} reaches test EM $0.874$ versus $0.875$ for \texttt{single\_small\_large\_semantic}$.$ At EM $\ge 0.86$, the required storage is $0.658$ versus $0.662$.

\subsection{Pooling and loss ablation}
Retaining the strongest feature candidates, we compared pooling strategies (\texttt{mean} vs.\ \texttt{last\_token}) and training objectives (MSE vs.\ Margin Ranking). 

\begin{figure}[ht]
  \centering
  \includegraphics[width=0.9\linewidth]{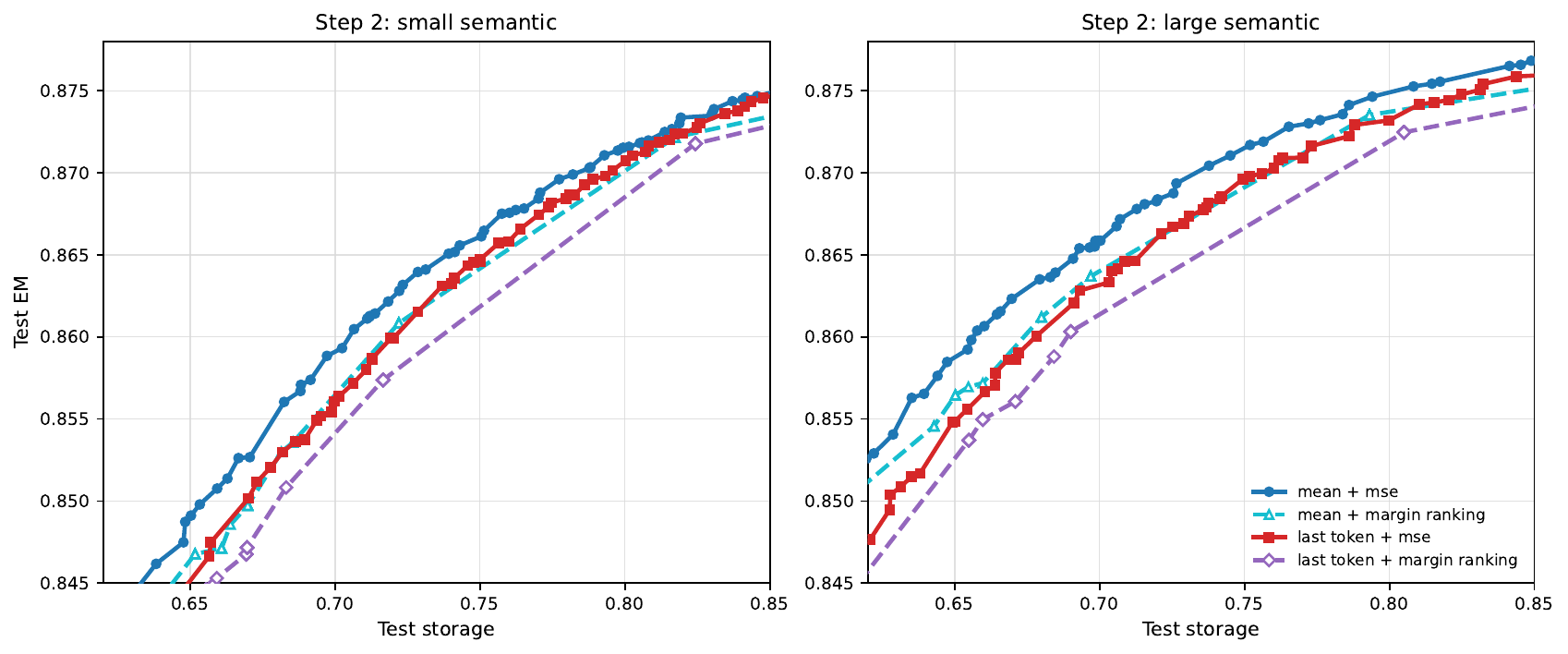}
  \caption{Step 2 pooling-and-loss ablation on the held-out test set.}
  \label{fig:ablation_step2}
\end{figure}

Figure~\ref{fig:ablation_step2} shows mean pooling with MSE gives the strongest overall frontier. On the large-semantic router, at storage $\le 0.8$, \texttt{mean+mse} reaches test EM $0.875$, compared with $0.874$ for \texttt{mean+margin\_ranking}, $0.873$ for \texttt{last\_token+mse}, and $0.860$ for \texttt{last\_token+margin\_ranking}. At EM $\ge 0.86$, it requires storage $0.658$, versus $0.680$, $0.678$, and $0.690$, respectively. 

\subsection{Escalation thresholds analysis}
\label{sec:appendix_cascade_numbers}
In Section \ref{subsec:ablations_summary}, we summarized the thresholded cascade's performance. Here we provide the precise ablation numbers supporting Figure~\ref{fig:main_test_frontier}. At storage caps 0.6, 0.7, and 0.8, the $\rho \le 0.6$ cascade frontier is only $0.0004$, $0.0022$, and $0.0010$ EM below the large router frontier, \textbf{while improving over the small router frontier by $0.0120$, $0.0048$, and $0.0021$, respectively}. Even under the stricter $\rho \le 0.4$ budget, the gap to the large router stays within $0.0031$ over the same range, proving that moderate escalation tightly bounds the theoretical optimum.

\subsection{Cascade gate design ablation}
\label{sec:appendix_gate_ablation}
To isolate the contribution of the escalation controller itself, we conducted a dedicated cascade-gate ablation. 

\begin{figure}[ht]
  \centering
  \includegraphics[width=0.9\linewidth]{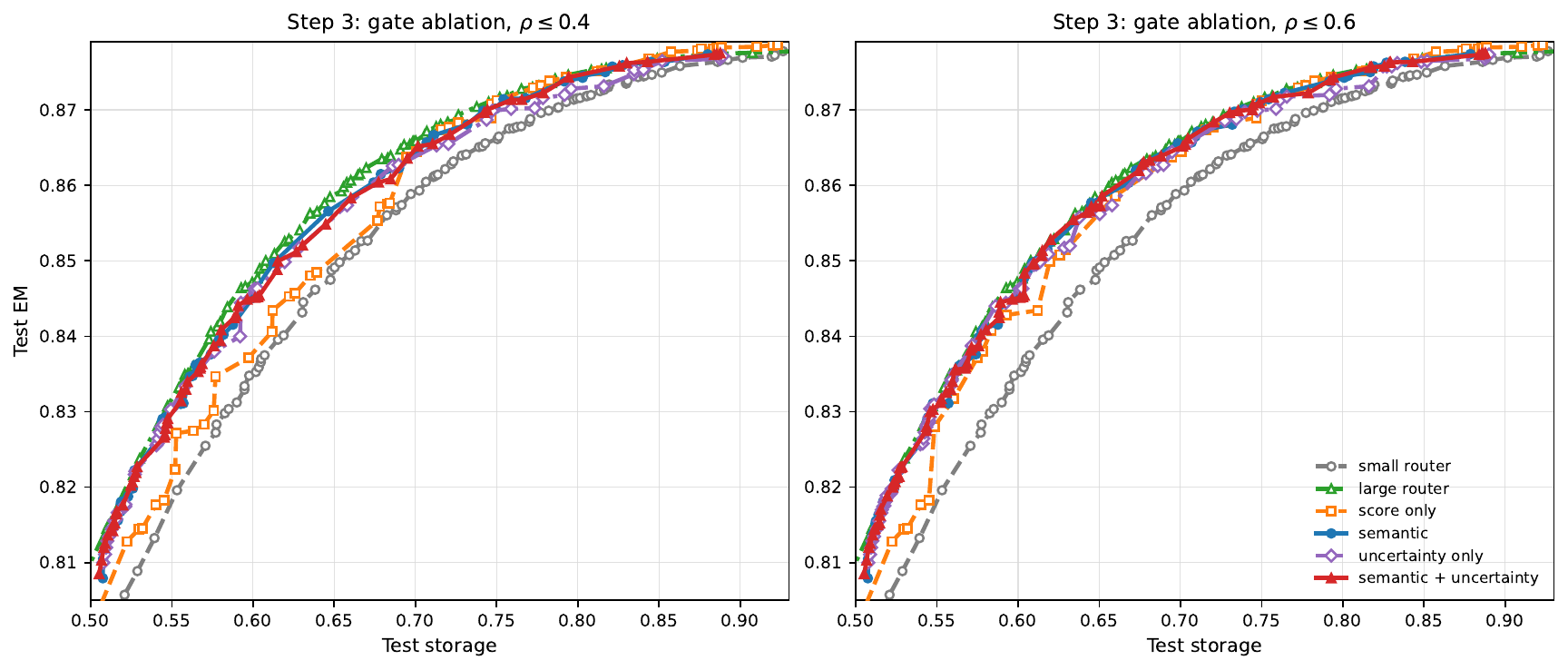}
  \caption{Step 3 cascade-gate ablation on the held-out test set under escalation thresholds $\rho \in \{0.4, 0.6\}$.}
  \label{fig:appendix_gate_ablation}
\end{figure}

Figure~\ref{fig:appendix_gate_ablation} shows that all learned gates close most of the gap between the small router and large router anchors. Under $\rho \le 0.6$, the \texttt{cascade\_gate\_semantic} gate reaches test EM $0.8451$ at storage $\le 0.6$ and $0.8656$ at storage $\le 0.7$, compared with $0.8428$ and $0.8638$ for \texttt{cascade\_gate\_score\_only}. The full \texttt{cascade\_gate\_semantic\_uncertainty} gate then recovers the strongest high-storage endpoint, reaching EM $0.8743$ at storage $\le 0.8$, only $0.0003$ below the large router frontier. Under the stricter $\rho \le 0.4$ budget, \texttt{cascade\_gate\_semantic\_uncertainty} also reaches EM $\ge 0.85$ at the lowest storage ($0.6267$), compared with $0.6464$ for \texttt{cascade\_gate\_semantic} and $0.6766$ for \texttt{cascade\_gate\_score\_only}. 

% \subsection{Write-back and consolidation analysis}
% In Section \ref{subsec:writeback_results}, we compared the behavioral labels of 30,203 held-out knowledge items before and after consolidation to better understand the epistemic shift at the fact level. Refer to Table~\ref{tab:transition_matrix} for a detailed breakdown of behavioral labels.
% \label{sec:SFT_analysis}
% \begin{table}[ht]
%   \caption{Post-write-back label transition matrix. While many facts become \textit{non-write}, 1,752 previously stable facts degrade, indicating parametric interference.}
%   \label{tab:transition_matrix}
%   \centering
%   \small
%   \begin{tabular}{lcccr}
%     \toprule
%     \textbf{Base \textbackslash{} SFT} & \textbf{non-write} & \textbf{write-new} & \textbf{write-update} & \textbf{Total (Base)} \\
%     \midrule
%     \textbf{non-write}    & 5,347  & 0 & 1,752 & 7,099 \\
%     \textbf{write-new}    & 5,785  & 0 & 7,939 & 13,724 \\
%     \textbf{write-update} & 4,898  & 0 & 4,482 & 9,380 \\
%     \midrule
%     \textbf{Total (SFT)}  & 16,030 & 0 & 14,173& \textbf{30,203} \\
%     \bottomrule
%   \end{tabular}
% \end{table}

\newpage
\section*{NeurIPS Paper Checklist}

\begin{enumerate}

\item {\bf Claims}
    \item[] Question: Do the main claims made in the abstract and introduction accurately reflect the paper's contributions and scope?
    \item[] Answer: \answerYes{} % Replace by \answerYes{}, \answerNo{}, or \answerNA{}.
    \item[] Justification: The main claims regarding the dual-layer memory architecture, cost-aware write routing, and slow consolidation are clearly stated in Section 1 and directly supported by the empirical evaluations in Section 4.

\item {\bf Limitations}
    \item[] Question: Does the paper discuss the limitations of the work performed by the authors?
    \item[] Answer: \answerYes{} % Replace by \answerYes{}, \answerNo{}, or \answerNA{}.
    \item[] Justification: The limitations, including the computational cost of offline SFT write-back and the lack of a principled mechanism to resolve temporally conflicting updates, are explicitly discussed in Section 5.

\item {\bf Theory assumptions and proofs}
    \item[] Question: For each theoretical result, does the paper provide the full set of assumptions and a complete (and correct) proof?
    \item[] Answer: \answerNA{} % Replace by \answerYes{}, \answerNo{}, or \answerNA{}.
    \item[] Justification: The paper is primarily empirical and systems-oriented; it does not present formal theoretical proofs or theorems.

    \item {\bf Experimental result reproducibility}
    \item[] Question: Does the paper fully disclose all the information needed to reproduce the main experimental results of the paper to the extent that it affects the main claims and/or conclusions of the paper (regardless of whether the code and data are provided or not)?
    \item[] Answer: \answerYes{} % Replace by \answerYes{}, \answerNo{}, or \answerNA{}.
    \item[] Justification: The full experimental setup, baseline definitions, hyperparameter trade-offs, routing methodology, and the deterministic dataset synthesis algorithms are detailed in Sections 3 and 4, as well as Appendices A-C.

\item {\bf Open access to data and code}
    \item[] Question: Does the paper provide open access to the data and code, with sufficient instructions to faithfully reproduce the main experimental results, as described in supplemental material?
    \item[] Answer: \answerNo{} % Replace by \answerYes{}, \answerNo{}, or \answerNA{}.
    \item[] Justification: As stated in the abstract, the code and synthesized dataset will be fully released upon acceptance. Detailed algorithmic descriptions and implementation protocols are provided in Section 3, Section 4, and Appendix A-C to facilitate reproduction.

\item {\bf Experimental setting/details}
    \item[] Question: Does the paper specify all the training and test details (e.g., data splits, hyperparameters, how they were chosen, type of optimizer) necessary to understand the results?
    \item[] Answer: \answerYes{} % Replace by \answerYes{}, \answerNo{}, or \answerNA{}.
    \item[] Justification: The training strategies, data splits, and evaluation metrics are detailed in Section 4 and Appendix A. The hyperparameter selection process (e.g., sweeping penalty terms $\lambda_s$, $\lambda_e$ on the validation split) and specific optimization details are explicitly documented in Appendix B and C.

\item {\bf Experiment statistical significance}
    \item[] Question: Does the paper report error bars suitably and correctly defined or other appropriate information about the statistical significance of the experiments?
    \item[] Answer: \answerNo{} % Replace by \answerYes{}, \answerNo{}, or \answerNA{}.
    \item[] Justification: Error bars are not reported because the online streaming benchmark evaluates deterministic routing policies over a fixed, large-scale dataset of 300 test episodes (150,000 queries total), making variance negligible.

\item {\bf Experiments compute resources}
    \item[] Question: For each experiment, does the paper provide sufficient information on the computer resources (type of compute workers, memory, time of execution) needed to reproduce the experiments?
    \item[] Answer: \answerYes{} % Replace by \answerYes{}, \answerNo{}, or \answerNA{}.
    \item[] Justification: We explicitly state the hardware compute resources (a node equipped with 4 NVIDIA H20 GPUs) used for the experiments in Section 4.2.
    
\item {\bf Code of ethics}
    \item[] Question: Does the research conducted in the paper conform, in every respect, with the NeurIPS Code of Ethics \url{https://neurips.cc/public/EthicsGuidelines}?
    \item[] Answer: \answerYes{} % Replace by \answerYes{}, \answerNo{}, or \answerNA{}.
    \item[] Justification: The research is strictly methodological and fully conforms to the NeurIPS Code of Ethics.

\item {\bf Broader impacts}
    \item[] Question: Does the paper discuss both potential positive societal impacts and negative societal impacts of the work performed?
    \item[] Answer: \answerNo{} % Replace by \answerYes{}, \answerNo{}, or \answerNA{}.
    \item[] Justification: The paper focuses on foundational agent memory optimization and architecture design. It does not introduce specific downstream societal risks beyond those inherent to general large language models.
    
\item {\bf Safeguards}
    \item[] Question: Does the paper describe safeguards that have been put in place for responsible release of data or models that have a high risk for misuse (e.g., pre-trained language models, image generators, or scraped datasets)?
    \item[] Answer: \answerNA{} % Replace by \answerYes{}, \answerNo{}, or \answerNA{}.
    \item[] Justification: We do not release any new pre-trained base models or scraped datasets that pose a high risk for misuse.

\item {\bf Licenses for existing assets}
    \item[] Question: Are the creators or original owners of assets (e.g., code, data, models), used in the paper, properly credited and are the license and terms of use explicitly mentioned and properly respected?
    \item[] Answer: \answerYes{} % Replace by \answerYes{}, \answerNo{}, or \answerNA{}.
    \item[] Justification: Existing assets, including the foundational ZsRE benchmark and the Qwen base models used for evaluation, are appropriately cited in the main text.

\item {\bf New assets}
    \item[] Question: Are new assets introduced in the paper well documented and is the documentation provided alongside the assets?
    \item[] Answer: \answerYes{} % Replace by \answerYes{}, \answerNo{}, or \answerNA{}.
    \item[] Justification: The paper introduces multiple new assets: (1) A novel streaming memory benchmark, whose synthesis algorithm (Algorithm 1), behavioral labeling taxonomy, and structural guarantees are thoroughly documented in Appendix A. (2) The codebase and algorithmic framework for the Dual-Layer Agentic Memory, which includes the implementations of the small-to-large cost-aware write router cascade, the escalation gating mechanism, and the periodic write-back (SFT) pipeline. Detailed mathematical formulations, training objectives, and deployment protocols for these algorithms are provided in Section 3 and Appendices B and C. As stated in the abstract, the complete benchmark datasets, evaluation environment, and framework source code will be publicly released upon acceptance to ensure full reproducibility.

\item {\bf Crowdsourcing and research with human subjects}
    \item[] Question: For crowdsourcing experiments and research with human subjects, does the paper include the full text of instructions given to participants and screenshots, if applicable, as well as details about compensation (if any)? 
    \item[] Answer: \answerNA{} % Replace by \answerYes{}, \answerNo{}, or \answerNA{}.
    \item[] Justification: The research does not involve crowdsourcing or human subjects.

\item {\bf Institutional review board (IRB) approvals or equivalent for research with human subjects}
    \item[] Question: Does the paper describe potential risks incurred by study participants, whether such risks were disclosed to the subjects, and whether Institutional Review Board (IRB) approvals (or an equivalent approval/review based on the requirements of your country or institution) were obtained?
    \item[] Answer: \answerNA{} % Replace by \answerYes{}, \answerNo{}, or \answerNA{}.
    \item[] Justification: The research does not involve human subjects.

\item {\bf Declaration of LLM usage}
    \item[] Question: Does the paper describe the usage of LLMs if it is an important, original, or non-standard component of the core methods in this research? Note that if the LLM is used only for writing, editing, or formatting purposes and does \emph{not} impact the core methodology, scientific rigor, or originality of the research, declaration is not required.
    %this research? 
    \item[] Answer: \answerNA{} % Replace by \answerYes{}, \answerNo{}, or \answerNA{}.
    \item[] Justification: LLMs were used solely for writing assistance and text editing (e.g., grammar, spelling, word choice), which does not impact the core methodology and does not require a formal declaration per NeurIPS guidelines.

\end{enumerate}

\end{document}